\documentclass[11pt]{article}

\usepackage[margin=1in]{geometry}
\usepackage{amsmath,amssymb}
\usepackage{booktabs}
\usepackage{graphicx}
\graphicspath{{figures/}{./}}
\usepackage{xcolor}
\usepackage{hyperref}
\usepackage{natbib}
\usepackage{authblk}

\hypersetup{colorlinks=true, linkcolor=blue, citecolor=blue, urlcolor=blue}

\title{Small Vision-Language Models Know When They Are Wrong\\
But Cannot Say So:\\
\large A Two-Model Study of Stated versus Internal Confidence\\
Under Realistic Image Degradation}

\author[1]{M M Asif Ferdous}
\affil[1]{Independent Researcher}

\date{\today}

\begin{document}
\maketitle

\begin{abstract}
Vision-language models (VLMs) are increasingly deployed on consumer hardware
where input images are degraded by compression, camera shake, and poor lighting.
In such settings a reliable uncertainty signal matters more than raw accuracy,
because it determines when a system should defer rather than answer. We evaluate
two small open-weight VLMs -- Qwen2-VL-2B-Instruct and SmolVLM-Instruct -- across
six realistic photographic degradations at three severity levels, comparing two
confidence signals: the confidence the model \emph{states} in natural language,
and the model's own mean token probability over its generated answer. Across
3{,}800 predictions we find a large and consistent gap. Verbalized confidence in
Qwen2-VL is almost constant (mean 0.87--0.90 across all conditions) and detects
its own errors at chance level (AUROC 0.39--0.75, typically $\approx$0.50),
while internal token probability from the same model separates correct from
incorrect answers with AUROC 0.92--0.99. In SmolVLM, verbalized confidence
proved largely unobtainable: across three prompt templates only one of five
pilot attempts produced a parseable confidence value, while internal probability
again yielded above-chance error detection (AUROC 0.54--0.92). Both models fail
in the same place: under severe underexposure, accuracy collapses
($0.99\rightarrow0.22$ for Qwen2-VL, $0.97\rightarrow0.42$ for SmolVLM) while
both confidence signals barely move, and internal error-detection falls to
chance. We conclude that small VLMs encode usable self-knowledge that their
verbalized output does not express, that internal probability is therefore the
better deferral signal in constrained deployment, and that neither signal should
be trusted under severe low-light conditions.
\end{abstract}

% ============================================================
\section{Introduction}
% ============================================================

The practical deployment of vision-language models increasingly happens at the
small end of the parameter range. Models of two to eight billion parameters run
on consumer GPUs and edge devices, and are chosen for cost, latency, privacy, or
the absence of reliable network connectivity. In these settings the images
reaching the model are rarely clean benchmark photographs: they are compressed
by messaging applications, blurred by handheld capture, underexposed indoors, or
washed out by direct light.

Accuracy alone is an insufficient description of how such a system behaves. What
determines whether a deployment is safe is whether the model's expressed
uncertainty tracks its actual reliability -- whether, when the model becomes
unreliable, anything in its output indicates this. A model that is wrong and
signals doubt can defer to a human. A model that is wrong and signals confidence
cannot.

Generative VLMs offer two distinct routes to an uncertainty estimate. The first
is to ask the model directly and read the number it states -- \emph{verbalized
confidence}. The second is to read the model's own probability distribution over
the tokens it generated -- \emph{internal confidence}. These are not obviously
the same quantity, and the relationship between them under input degradation has
not been systematically examined for small open-weight models.

This paper asks three questions:

\begin{itemize}
\item \textbf{RQ1.} How does the accuracy of small VLMs degrade under realistic
photographic corruption?
\item \textbf{RQ2.} Does verbalized confidence track that degradation?
\item \textbf{RQ3.} Does internal token probability track it better?
\end{itemize}

Our contributions are: (i) a calibration-centred benchmark of two small
open-weight VLMs across six realistic degradations at graded severities, with
bootstrap confidence intervals; (ii) evidence that verbalized and internal
confidence diverge sharply, with internal probability providing near-ceiling
error detection where verbalized confidence performs at chance; and (iii) an
honest characterisation of where both signals fail.

% ============================================================
%  Related Work
%  Positioning verified against arXiv:2504.03440
%  (Borszukovszki, de Jong, Valdenegro-Toro 2025)
% ============================================================

\section{Related Work}

\subsection{Verbalized uncertainty in language and vision-language models}

Because generative models emit text rather than a probability vector, a natural
approach to uncertainty estimation is to prompt the model to state its own
confidence. \citet{xiong2024} evaluated confidence-elicitation strategies in
LLMs and reported that models are overconfident, with the majority of confidence
scores falling in the 80--100 range. \citet{tian2023} found that for models
fine-tuned with RLHF, verbalized confidence can be better calibrated than
sampling-based estimates of internal token probabilities -- a result that has
made verbalized uncertainty the default choice in much subsequent VLM work.
\citet{groot2024} extended verbalized confidence estimation to visual
question answering and found VLMs poorly calibrated and severely overconfident.

\subsection{Verbalized uncertainty under image corruption}

The work closest to ours is \citet{borszukovszki2025}, who evaluated three
proprietary VLMs (GPT-4 Vision, Gemini Pro Vision, Claude 3 Opus) on three
corruption types (Gaussian noise, defocus blur, JPEG compression) at five
severity levels, across easy VQA, hard VQA, and counting tasks. They report
that increasing corruption severity increases Expected Calibration Error, that
models remain overconfident throughout, and that higher refusal rates improve
calibration.

We replicate their central findings on open-weight models: our Qwen2-VL-2B
states a near-constant confidence of 0.87--0.90 regardless of accuracy,
consistent with the 80--100 clustering they and \citet{xiong2024} report, and
calibration degrades sharply under severe corruption.

Our contribution is orthogonal to theirs, and their paper states precisely why.
They motivate their use of verbalized confidence by noting that for the models
they study, ``since these models are proprietary, we don't have access to these
individual token probabilities,'' and describe their contribution as extending
the research ``into VLMs where internal token probabilities are not
available.'' Their design therefore \emph{cannot} compare verbalized confidence
against the model's own token-level distribution -- the signal is inaccessible by
construction.

By evaluating open-weight models, we recover exactly that signal. This allows
the comparison their setting excludes: verbalized against internal confidence,
computed on the same predictions, under the same corruptions. Our central
finding -- that internal token probability detects errors with AUROC up to 0.99
while verbalized confidence from the same model performs at chance -- is only
measurable in the open-weight setting.

We differ in three further respects. First, \citet{borszukovszki2025} evaluate
frontier proprietary models; we evaluate small ($\approx$2B) open-weight models
representative of constrained deployment. Second, they use three synthetic
corruption families from \citet{michaelis2019}; we use six degradations selected
to approximate real phone-camera artifacts, including low light and glare, and
find that low light produces a failure mode not visible in their corruption set.
Third, they report accuracy, confidence, and ECE; we add error-detection AUROC
with bootstrap confidence intervals, which we argue is the more diagnostic
metric when a confidence signal has near-zero variance
(Section~\ref{sec:results}).

Finally, \citet{borszukovszki2025} explicitly identify temperature scaling as
future work, noting it ``would be interesting to explore if this overconfidence
in VLMs could be treated with temperature scaling.'' We test this directly and
report a negative result: post-hoc temperature scaling improves ECE under mild
degradation but \emph{worsens} it under severe low light, because rescaling a
near-constant signal cannot recover information the signal never contained.

\subsection{Corruption robustness and calibration under shift}

The systematic study of corruption robustness was formalised by
\citet{hendrycks2019}, who introduced parameterised corruption families at
calibrated severity levels; \citet{michaelis2019} extended this taxonomy. We
follow their methodological template of graded severities reported as a function
of severity rather than in aggregate.

Separately, \citet{ovadia2019} established that predictive uncertainty degrades
under dataset shift, and \citet{guo2017} showed that post-hoc temperature scaling
corrects overconfidence in classifiers. Our negative temperature-scaling result
is consistent with the mechanism \citet{guo2017} describe: temperature scaling
redistributes probability mass but cannot create discriminative signal where
none exists.

\subsection{Positioning summary}

To our knowledge, no prior work compares verbalized against internal confidence
on the same predictions in vision-language models under image degradation. The
closest work \citep{borszukovszki2025} identifies the absence of internal token
probabilities as a defining constraint of its setting. This paper occupies that
gap.

% ============================================================
\section{Method}
% ============================================================

\subsection{Models}

We evaluate two small open-weight instruction-tuned VLMs:
\textbf{Qwen2-VL-2B-Instruct} \citep{wang2024qwen2vl} and
\textbf{SmolVLM-Instruct} \citep{smolvlm}. Both run in fp16 on a single free-tier NVIDIA T4
(16\,GB). Generation is greedy (\texttt{do\_sample=False}) throughout for
reproducibility.

\subsection{Dataset}

A 100-item subset of Food101 \citep{bossard2014food101} (test split, streamed, seed 0), posed as four-option
multiple choice: one correct label and three distractors sampled uniformly from
the remaining 100 classes. Food101 provides real photographs at usable
resolution, appropriate for a corruption study.

\subsection{Degradations}

Six degradation families were chosen to approximate real phone-camera artifacts,
each applied at three severity levels (Table~\ref{tab:degradations}).

\begin{table}[h]
\centering
\caption{Degradation families and severity parameters.}
\label{tab:degradations}
\begin{tabular}{llc}
\toprule
Family & Mechanism & Severities (1/2/3) \\
\midrule
JPEG compression & re-encode at low quality & $q = 30$ / 15 / 7 \\
Motion blur & horizontal box kernel & $r = 2$ / 4 / 7 \\
Low light & brightness reduction $+$ Gaussian noise & $\times$0.5 / 0.3 / 0.15 \\
Glare & brightness amplification & $\times$1.6 / 2.2 / 3.0 \\
Rotation & in-plane tilt, black fill & $5^\circ$ / $12^\circ$ / $20^\circ$ \\
Resample & downscale then upscale & $\times$0.5 / 0.3 / 0.15 \\
\bottomrule
\end{tabular}
\end{table}

With the clean baseline this yields 19 conditions per model, 100 items each:
1{,}900 predictions per model and 3{,}800 in total.

\subsection{Confidence signals}

\paragraph{Verbalized confidence.} The model is prompted to answer and state an
integer confidence in $[0,100]$, normalised to $[0,1]$. The prompt template was
selected empirically: three candidates were piloted on five items and the one
with the highest parse rate retained. For Qwen2-VL a few-shot template achieved
5/5 parseable outputs; for SmolVLM the best of three achieved only 1/5
(Section~\ref{sec:unobtainable}). Exact templates appear in
Appendix~\ref{app:prompts}.

\paragraph{Internal confidence.} During generation we retain per-step output
distributions and compute the mean probability assigned to each token actually
emitted in the answer span:
\begin{equation}
c_{\text{int}} = \frac{1}{T}\sum_{t=1}^{T} p_\theta(y_t \mid y_{<t}, x),
\end{equation}
where $y_t$ is the $t$-th generated token and $x$ the multimodal input. This is
continuous in $(0,1]$ and requires no additional forward passes.

\subsection{Answer matching}

Generated answers rarely match gold labels exactly. We normalise case and
punctuation, accept exact matches, and fall back to unique containment (the
prediction contains exactly one option string). The match method is recorded per
prediction for transparency.

\subsection{Metrics}

We report accuracy, Expected Calibration Error \citep{naeini2015} with 10
bins, Brier score \citep{brier1950}, and
\textbf{error-detection AUROC} -- the probability that a randomly chosen correct
answer receives higher confidence than a randomly chosen incorrect one. AUROC is
our primary metric because it directly measures whether a confidence signal is
usable for deferral. Confidence intervals are percentile bootstrap with
$B = 2000$ resamples.

% ============================================================
\section{Results}
\label{sec:results}
% ============================================================

\subsection{Accuracy is robust except under severe low light}

Qwen2-VL achieves 0.99 clean accuracy and remains at 0.87--1.00 across five of
six degradation families at all severities. The exception is low light:
$0.99 \rightarrow 0.92$ (s2) $\rightarrow$ \textbf{0.22} (s3), approaching the
0.25 chance rate for four-option multiple choice. SmolVLM shows the same pattern
at slightly lower absolute accuracy: 0.97 clean, 0.74--0.97 across most
conditions, collapsing to \textbf{0.42} at low light s3.

\begin{figure}[h]
\centering
\includegraphics[width=\textwidth]{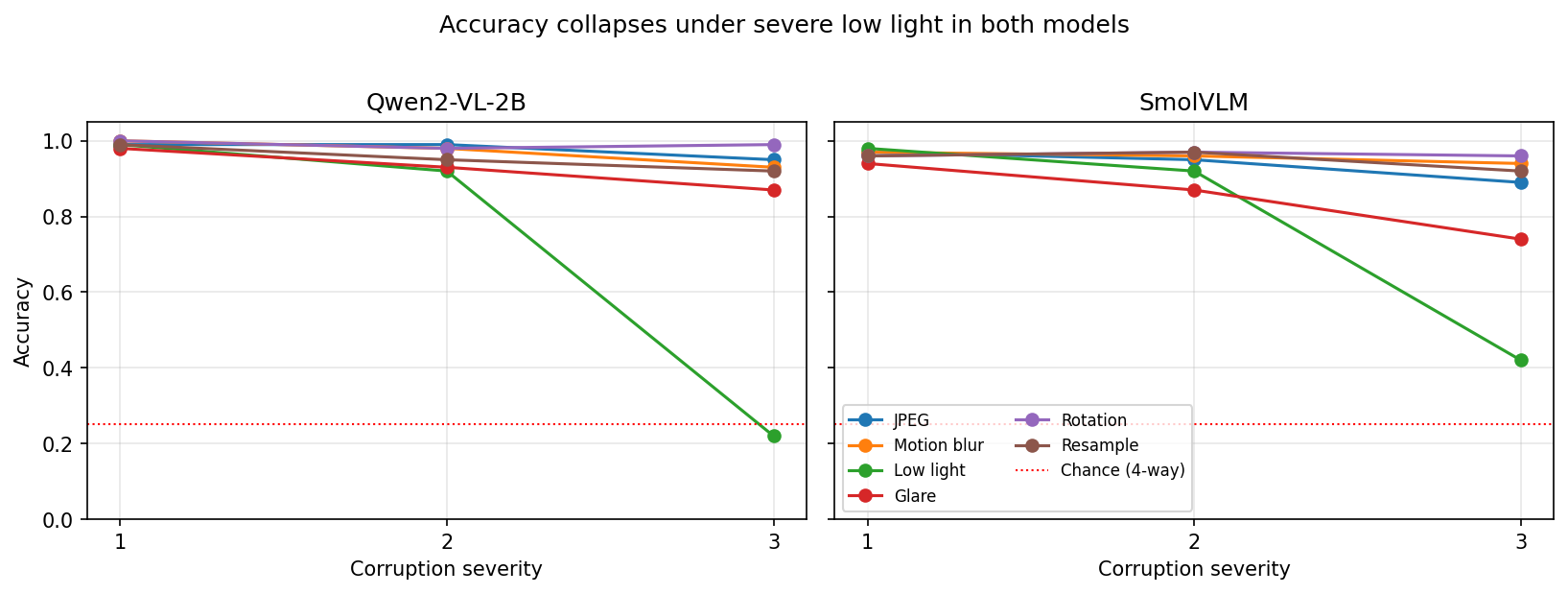}
\caption{Accuracy as a function of corruption severity for both models. Five of
six degradation families produce little accuracy loss; low light (green)
collapses toward the four-way chance rate (dotted red) in both models. The
replication across two independent model families is the point of this figure.}
\label{fig:accuracy}
\end{figure}

\paragraph{Finding 1.} \emph{Accuracy in small VLMs is largely insensitive to
compression, blur, glare, tilt and resampling, but collapses under severe
underexposure. This replicates across both models.}

\subsection{Verbalized confidence is constant and uninformative}

For Qwen2-VL, mean verbalized confidence ranges from 0.87 to 0.90 across all 19
conditions. It does not respond to severity, degradation family, or accuracy. At
low light s3 -- accuracy 0.22 -- the model still states 0.87.
Table~\ref{tab:main} gives error-detection AUROC with 95\% bootstrap intervals.

\paragraph{Finding 2.} \emph{Verbalized confidence in Qwen2-VL is effectively a
constant and carries almost no information about correctness.}

\subsection{Internal confidence detects errors}

\begin{table}[h]
\centering
\caption{Qwen2-VL-2B: error-detection AUROC for verbalized versus internal
confidence, with 95\% bootstrap confidence intervals ($B=2000$, $n\approx100$
per condition).}
\label{tab:main}
\begin{tabular}{lccc}
\toprule
Condition & Accuracy & AUROC (verbalized) & AUROC (internal) \\
\midrule
clean            & 0.99 [0.97, 1.00] & 0.49 [0.47, 0.51] & \textbf{0.92 [0.86, 0.97]} \\
glare s3         & 0.87 [0.80, 0.93] & 0.69 [0.55, 0.83] & \textbf{0.91 [0.80, 0.98]} \\
jpeg s3          & 0.95 [0.90, 0.99] & 0.88 [0.50, 1.00] & \textbf{0.96 [0.91, 1.00]} \\
low\_light s1    & 0.99 [0.97, 1.00] & 0.39 [0.35, 0.43] & \textbf{0.96 [0.92, 0.99]} \\
low\_light s2    & 0.92 [0.86, 0.97] & 0.61 [0.43, 0.79] & \textbf{0.92 [0.83, 0.99]} \\
low\_light s3    & 0.22 [0.14, 0.30] & 0.68 [0.54, 0.80] & 0.62 [0.48, 0.75] \\
motion\_blur s3  & 0.93 [0.88, 0.98] & 0.57 [0.50, 0.75] & \textbf{0.96 [0.90, 1.00]} \\
resample s3      & 0.92 [0.86, 0.97] & 0.50 [0.30, 0.70] & \textbf{0.95 [0.88, 1.00]} \\
rotation s3      & 0.99 [0.97, 1.00] & 0.50 [0.50, 0.50] & \textbf{0.98 [0.95, 1.00]} \\
\bottomrule
\end{tabular}
\end{table}

In clean, low\_light s1, resample s1 and rotation s3 the two intervals do not
overlap: the difference is statistically robust at this sample size.

\begin{figure}[h]
\centering
\includegraphics[width=\textwidth]{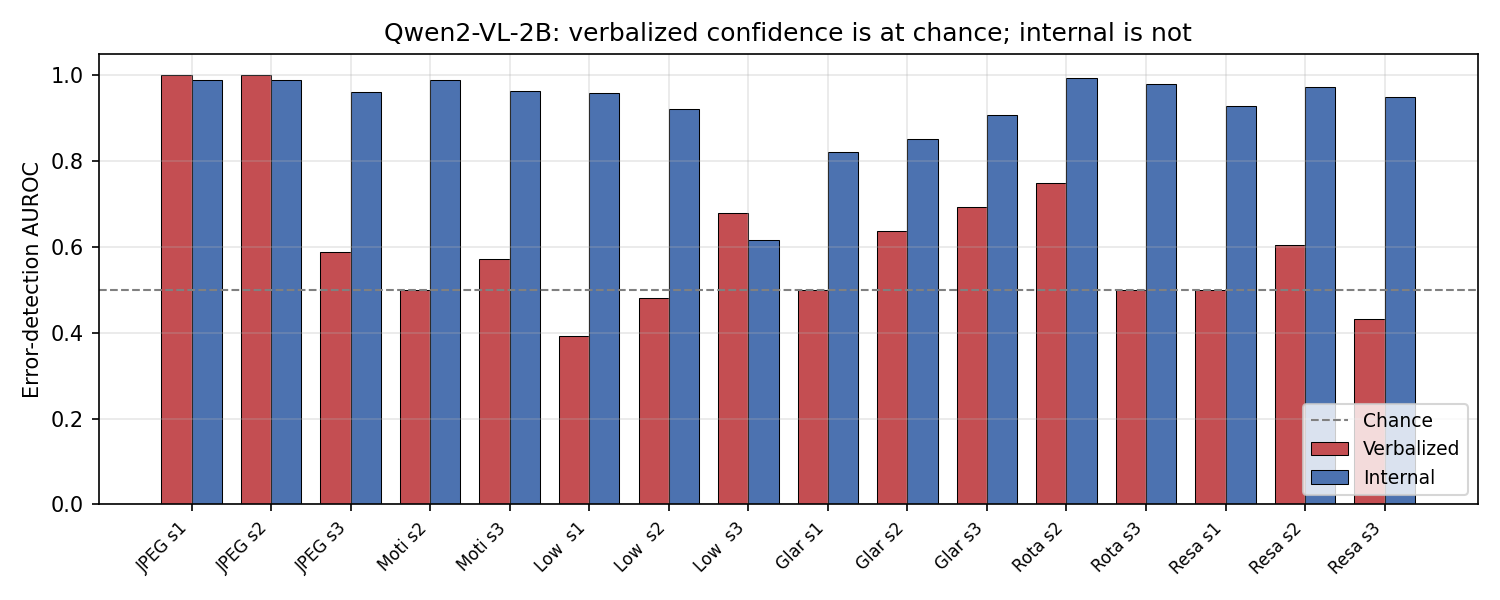}
\caption{Error-detection AUROC for verbalized (red) versus internal (blue)
confidence in Qwen2-VL-2B, across all conditions where AUROC is defined. Dashed
line marks chance (0.5). Verbalized confidence sits at or near chance in most
conditions and falls \emph{below} it under low light s1; internal confidence
from the same predictions ranges 0.82--0.99. The two JPEG conditions where
verbalized AUROC reaches 1.00 have near-perfect accuracy and correspondingly few
errors, so those values rest on a handful of points and should not be read as
evidence that verbalized confidence works there.}
\label{fig:headline}
\end{figure}

\paragraph{Finding 3 (headline).} \emph{The same model that cannot state a
useful confidence number nonetheless encodes a strong internal signal of its own
correctness.}

\subsection{The finding replicates in SmolVLM, more weakly}

SmolVLM internal AUROC is above 0.5 in all 19 conditions, ranging 0.54--0.96:
clean 0.77 [0.58, 1.00], low\_light s1 0.96 [0.92, 1.00], rotation s2 0.92
[0.82, 1.00], resample s1 0.81 [0.60, 0.98]. The direction replicates but the
effect is weaker and noisier than in Qwen2-VL. Several SmolVLM intervals include
or approach 0.5 -- glare s1 0.63 [0.32, 0.94], jpeg s1 0.62 [0.33, 0.98] -- and
are therefore \textbf{not distinguishable from chance at this sample size}. We
report these as inconclusive rather than positive.

\paragraph{Finding 4.} \emph{Internal confidence provides above-chance error
detection in both models, but its strength is model-dependent: strong and
consistent in Qwen2-VL (0.92--0.99), weaker and more variable in SmolVLM
(0.54--0.92).}

\subsection{Both signals fail under severe low light}

\begin{table}[h]
\centering
\caption{Both models fail together under severe underexposure.}
\label{tab:lowlight}
\begin{tabular}{lccc}
\toprule
Model & Clean acc. & Low light s3 acc. & Internal AUROC at s3 \\
\midrule
Qwen2-VL-2B & 0.99 & 0.22 & 0.62 [0.48, 0.75] \\
SmolVLM     & 0.97 & 0.42 & 0.54 [0.42, 0.66] \\
\bottomrule
\end{tabular}
\end{table}

For Qwen2-VL, accuracy falls 0.770 while verbalized confidence falls 0.029 and
internal confidence falls 0.057. Internal confidence is roughly twice as
responsive, but both are negligible against the accuracy drop. Both models'
internal AUROC intervals include 0.5 at this severity.

\begin{figure}[h]
\centering
\includegraphics[width=0.72\textwidth]{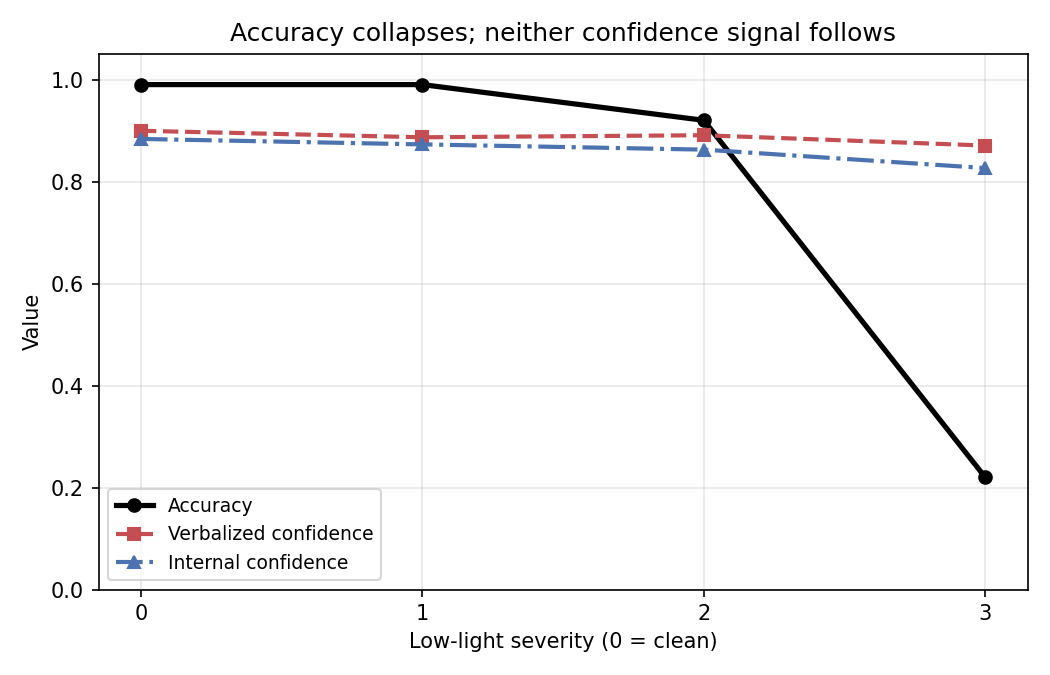}
\caption{Qwen2-VL-2B under increasing low light. Accuracy (black) falls from
0.99 to 0.22 while verbalized confidence (red) moves 0.029 and internal
confidence (blue) moves 0.057. Internal confidence is the more responsive of the
two, but the gap that opens between accuracy and both signals is the failure
mode this paper cautions against.}
\label{fig:divergence}
\end{figure}

\paragraph{Finding 5.} \emph{Neither signal tracks catastrophic failure.
Internal confidence is a useful deferral signal under mild-to-moderate
degradation and loses that property precisely under the severe degradation where
it would matter most. This replicates across both models.}

\subsection{Calibration error hides the difference between the two signals}

Expected Calibration Error is low for both signals in most conditions (roughly
0.02--0.12), which follows mechanically from the model being both accurate and
confident nearly everywhere. ECE rises sharply only at low light s3: 0.650 for
verbalized and 0.606 for internal confidence.

\begin{figure}[h]
\centering
\includegraphics[width=0.85\textwidth]{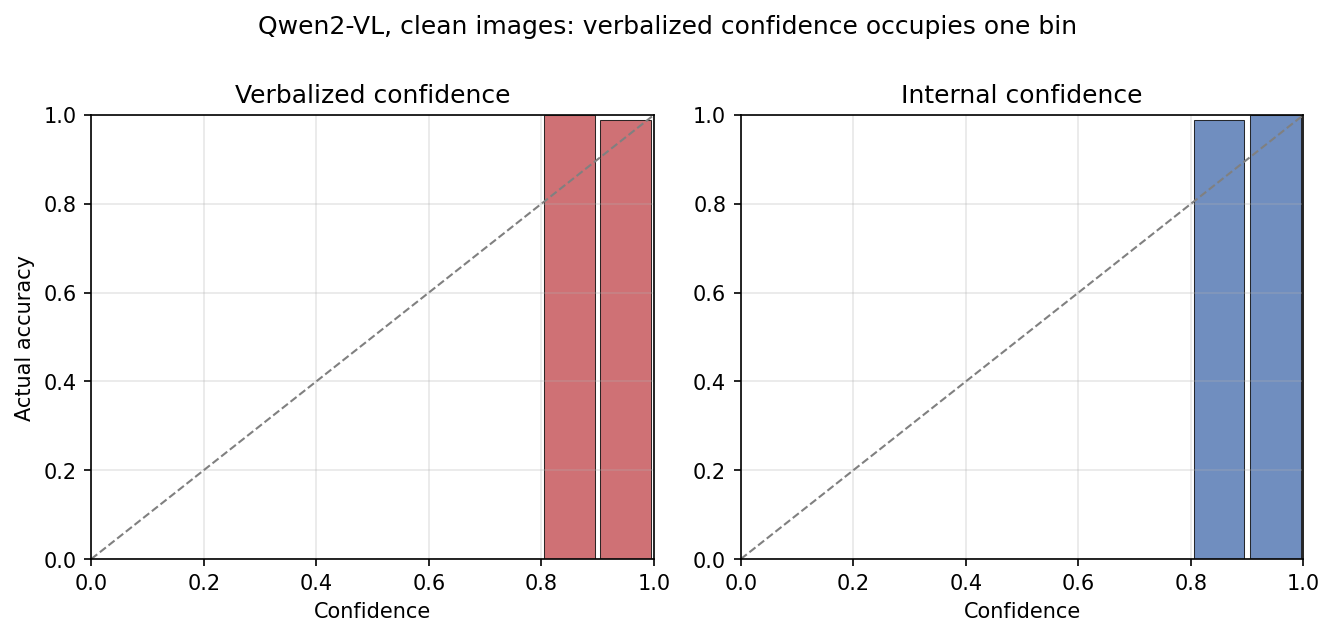}
\caption{Reliability diagrams for both confidence signals on clean images
(Qwen2-VL-2B). Both distributions are concentrated in $[0.8, 1.0]$ and both sit
near the diagonal, because the model is 99\% accurate here and both signals are
correspondingly high. \textbf{The reliability diagram cannot distinguish the two
signals}, even though their error-detection AUROC differs by more than 0.4
(0.49 versus 0.92). We include this as a methodological caution: calibration-error
measures assess whether confidence \emph{magnitude} matches accuracy, not whether
confidence \emph{discriminates} correct from incorrect answers. A signal with
near-zero variance can appear well calibrated while carrying no usable
information.}
\label{fig:reliability}
\end{figure}

We report ECE for comparability with prior VLM calibration work, but in this
setting it is the \emph{less} informative metric. Because verbalized confidence
is nearly constant, its ECE is dominated by the accuracy of the condition rather
than by any property of the confidence signal itself.
Figure~\ref{fig:reliability} makes this concrete: on clean images the two
signals are visually indistinguishable in a reliability diagram, despite an
error-detection AUROC gap of more than 0.4 on the very same predictions.

Error-detection AUROC is the more diagnostic measure in this regime, and we
recommend it for future work on verbalized confidence in small VLMs.

\subsection{Post-hoc temperature scaling}

In a single-signal run we fitted a temperature parameter on a held-out clean
split ($T = 0.50$) and applied it unchanged to all degraded conditions. ECE
improved substantially in most conditions (clean $0.090 \rightarrow 0.002$;
jpeg s1 $0.107 \rightarrow 0.002$; low\_light s1 $0.103 \rightarrow 0.007$;
rotation s3 $0.090 \rightarrow 0.002$) but \textbf{worsened} at low light s3,
$0.650 \rightarrow 0.756$. The interpretation is mechanical: rescaling a
near-constant signal redistributes a fixed value and cannot recover information
the signal never contained.

% ============================================================
\section{Discussion}
% ============================================================

\subsection{Why the two signals diverge}

Verbalized confidence requires the model to introspect and then serialise that
introspection into a number. This is a learned surface behaviour, and in small
instruction-tuned models it appears to collapse to a stock response -- 0.90 in
almost every case for Qwen2-VL. Internal token probability requires no
introspection: it is a direct read of the model's own distribution and cannot be
produced by imitation of a training pattern in the same way. The practical
implication is that the model has more self-knowledge than its text output
reveals.

\subsection{What practitioners should do}

For deployment on small open-weight VLMs, \textbf{do not use stated confidence
for deferral}. It is at chance in one of our models and largely unobtainable in
the other. Use mean token probability instead: it is free to compute, requires
no retraining, and provides usable error detection under mild-to-moderate
degradation. However, \textbf{do not treat it as a safety guarantee under low
light}. Both models retained high internal confidence while accuracy collapsed.
A deployment that may encounter severe underexposure needs an explicit
image-quality check upstream of the model, not a confidence threshold downstream
of it.

\subsection{Verbalized confidence may be unobtainable, not merely poor}
\label{sec:unobtainable}

SmolVLM did not simply give poor confidence numbers -- it largely refused to give
them. Across three prompt templates in a five-item pilot, only one attempt
produced a parseable value; the model returned bare answers (``Takoyaki.'',
``Pho.'') regardless of instruction. We report this as a qualitative observation
rather than a measured rate, and note it as a limitation: a more extensive
prompt search might succeed where ours did not.

\subsection{Limitations}

\begin{enumerate}
\item \textbf{Undefined AUROC cells.} Conditions at or near 1.00 accuracy
(motion\_blur s1, rotation s1 for Qwen2-VL) contain too few errors for AUROC to
be defined. These are reported as n/a and excluded from aggregates. They are
\emph{not} evidence of good calibration.
\item \textbf{Sample size.} $n \approx 100$ per condition. Several SmolVLM
intervals span 0.3--0.4 in width and are uninformative. A larger $n$ is the
single highest-value extension of this work.
\item \textbf{One dataset.} All results are on Food101 posed as four-way
multiple choice. Generality to open-ended VQA and other domains is untested.
\item \textbf{Prompt sensitivity.} Verbalized confidence was elicited with one
template per model, chosen by pilot. The near-constant 0.90 in Qwen2-VL may be
partly template-induced. An elicitation ablation is required before the claim
can be stated in full generality.
\item \textbf{Synthetic degradations.} Our corruptions approximate phone-camera
artifacts programmatically; they are not photographs actually taken under those
conditions.
\item \textbf{Two models, one size class.} Both models are $\approx$2B
parameters. Whether the verbalized/internal gap narrows with scale is untested
and is the most interesting open question this work raises.
\item \textbf{No refusal behaviour.} Our multiple-choice format forces a
response and therefore cannot capture refusal, which \citet{borszukovszki2025}
identify as a meaningful calibration mechanism: in their open-ended setting,
models that declined to answer under severe corruption achieved better
calibration than those that guessed. Whether small open-weight models refuse
under degradation, and whether refusal correlates with low internal confidence,
is an open question our design cannot address.
\end{enumerate}

% ============================================================
\section{Conclusion}
% ============================================================

Across 3{,}800 predictions from two small open-weight vision-language models
under six realistic photographic degradations, we find that verbalized
confidence is close to useless -- constant and chance-level in Qwen2-VL-2B,
largely unobtainable in SmolVLM -- while internal token probability from the same
predictions detects errors with AUROC up to 0.99. Small VLMs appear to encode
substantially more self-knowledge than their natural-language output expresses.
Both models nonetheless fail together under severe underexposure, where accuracy
collapses and neither signal responds. We recommend internal token probability
over stated confidence as a deferral signal in constrained deployment, coupled
with an explicit upstream image-quality check.

% ============================================================
\section{Reproducibility}
% ============================================================

All code, configurations, per-prediction CSVs, and figures are available at
\url{https://github.com/Asif-Ferdous/vlm-reliability}. Experiments run on a single free-tier NVIDIA T4; the
full study reproduces in approximately 90 minutes of GPU time. Random seeds are
fixed throughout.

% ============================================================
\bibliographystyle{plainnat}
\bibliography{references}

\appendix

\section{Full results tables}
Tables~\ref{tab:appendix-qwen} and \ref{tab:appendix-smol} give the complete
per-condition results for both models. Values are transcribed from
\texttt{summary\_with\_CIs.csv} and \texttt{summary\_smolvlm.csv}; brackets give
95\% percentile bootstrap intervals ($B=2000$). Cells marked n/a have
insufficient errors for AUROC to be defined.

\begin{table}[h]
\centering
\small
\caption{Qwen2-VL-2B-Instruct: complete results, all 19 conditions.}
\label{tab:appendix-qwen}
\begin{tabular}{lccc}
\toprule
Condition & Accuracy & AUROC (verbalized) & AUROC (internal) \\
\midrule
clean            & 0.99 [0.97, 1.00] & 0.49 [0.47, 0.51] & 0.92 [0.86, 0.97] \\
glare s1         & 0.98 [0.95, 1.00] & 0.50 [0.50, 0.50] & 0.82 [0.67, 0.95] \\
glare s2         & 0.93 [0.88, 0.97] & 0.64 [0.49, 0.83] & 0.85 [0.72, 0.96] \\
glare s3         & 0.87 [0.80, 0.93] & 0.69 [0.55, 0.83] & 0.91 [0.80, 0.98] \\
jpeg s1          & 0.99 [0.97, 1.00] & 1.00 [1.00, 1.00] & 0.99 [0.97, 1.00] \\
jpeg s2          & 0.99 [0.97, 1.00] & 1.00 [1.00, 1.00] & 0.99 [0.97, 1.00] \\
jpeg s3          & 0.95 [0.90, 0.99] & 0.88 [0.50, 1.00] & 0.96 [0.91, 1.00] \\
low\_light s1    & 0.99 [0.97, 1.00] & 0.39 [0.35, 0.43] & 0.96 [0.92, 0.99] \\
low\_light s2    & 0.92 [0.86, 0.97] & 0.61 [0.43, 0.79] & 0.92 [0.83, 0.99] \\
low\_light s3    & 0.22 [0.14, 0.30] & 0.68 [0.54, 0.80] & 0.62 [0.48, 0.75] \\
motion\_blur s1  & 1.00 [1.00, 1.00] & n/a               & n/a               \\
motion\_blur s2  & 0.98 [0.95, 1.00] & 1.00 [1.00, 1.00] & 0.99 [0.96, 1.00] \\
motion\_blur s3  & 0.93 [0.88, 0.98] & 0.57 [0.50, 0.75] & 0.96 [0.90, 1.00] \\
resample s1      & 0.99 [0.97, 1.00] & 0.50 [0.50, 0.50] & 0.93 [0.88, 0.97] \\
resample s2      & 0.95 [0.91, 0.99] & 0.60 [0.50, 0.84] & 0.97 [0.94, 1.00] \\
resample s3      & 0.92 [0.86, 0.97] & 0.50 [0.30, 0.70] & 0.95 [0.88, 1.00] \\
rotation s1      & 1.00 [1.00, 1.00] & n/a               & n/a               \\
rotation s2      & 0.98 [0.95, 1.00] & 0.75 [0.50, 1.00] & 0.99 [0.97, 1.00] \\
rotation s3      & 0.99 [0.97, 1.00] & 0.50 [0.50, 0.50] & 0.98 [0.95, 1.00] \\
\bottomrule
\end{tabular}
\end{table}

\begin{table}[h]
\centering
\small
\caption{SmolVLM-Instruct: complete results, all 19 conditions. Verbalized
confidence was not reliably obtainable (Section~\ref{sec:unobtainable}), so only
internal confidence is reported.}
\label{tab:appendix-smol}
\begin{tabular}{lccc}
\toprule
Condition & Accuracy & AUROC (internal) & Mean internal conf. \\
\midrule
clean            & 0.97 [0.93, 1.00] & 0.77 [0.58, 1.00] & 0.935 \\
glare s1         & 0.94 [0.89, 0.98] & 0.63 [0.32, 0.94] & 0.935 \\
glare s2         & 0.87 [0.80, 0.93] & 0.77 [0.60, 0.90] & 0.915 \\
glare s3         & 0.74 [0.65, 0.82] & 0.67 [0.55, 0.78] & 0.905 \\
jpeg s1          & 0.97 [0.93, 1.00] & 0.62 [0.33, 0.98] & 0.922 \\
jpeg s2          & 0.95 [0.90, 0.99] & 0.65 [0.35, 0.94] & 0.917 \\
jpeg s3          & 0.89 [0.83, 0.95] & 0.75 [0.58, 0.90] & 0.901 \\
low\_light s1    & 0.98 [0.95, 1.00] & 0.96 [0.92, 1.00] & 0.940 \\
low\_light s2    & 0.92 [0.86, 0.97] & 0.76 [0.55, 0.92] & 0.926 \\
low\_light s3    & 0.42 [0.32, 0.52] & 0.54 [0.42, 0.66] & 0.838 \\
motion\_blur s1  & 0.97 [0.93, 1.00] & 0.77 [0.48, 1.00] & 0.929 \\
motion\_blur s2  & 0.96 [0.92, 0.99] & 0.68 [0.42, 1.00] & 0.922 \\
motion\_blur s3  & 0.94 [0.89, 0.98] & 0.68 [0.40, 0.94] & 0.908 \\
resample s1      & 0.96 [0.92, 0.99] & 0.81 [0.60, 0.98] & 0.930 \\
resample s2      & 0.97 [0.93, 1.00] & 0.63 [0.35, 0.86] & 0.916 \\
resample s3      & 0.92 [0.86, 0.97] & 0.62 [0.50, 0.75] & 0.901 \\
rotation s1      & 0.96 [0.92, 0.99] & 0.91 [0.83, 1.00] & 0.943 \\
rotation s2      & 0.97 [0.93, 1.00] & 0.92 [0.82, 1.00] & 0.938 \\
rotation s3      & 0.96 [0.92, 0.99] & 0.92 [0.71, 1.00] & 0.935 \\
\bottomrule
\end{tabular}
\end{table}

\clearpage

\begin{table}[h]
\centering
\small
\caption{Post-hoc temperature scaling on Qwen2-VL verbalized confidence.
A single temperature $T=0.50$ was fitted on a held-out clean split and applied
unchanged to all degraded conditions. Note the failure at low light s3.}
\label{tab:appendix-temp}
\begin{tabular}{lcccc}
\toprule
Condition & ECE before & ECE after & Brier before & Brier after \\
\midrule
clean            & 0.090 & \textbf{0.002} & 0.018 & 0.010 \\
glare s1         & 0.080 & \textbf{0.008} & 0.026 & 0.020 \\
glare s2         & 0.061 & 0.057 & 0.063 & 0.067 \\
glare s3         & 0.055 & 0.116 & 0.106 & 0.123 \\
jpeg s1          & 0.107 & \textbf{0.002} & 0.017 & 0.010 \\
jpeg s2          & 0.107 & \textbf{0.002} & 0.017 & 0.010 \\
jpeg s3          & 0.112 & \textbf{0.028} & 0.039 & 0.038 \\
low\_light s1    & 0.103 & \textbf{0.007} & 0.022 & 0.010 \\
low\_light s2    & 0.039 & 0.067 & 0.075 & 0.080 \\
low\_light s3    & 0.650 & \emph{0.756} & 0.590 & 0.742 \\
motion\_blur s1  & 0.099 & \textbf{0.012} & 0.010 & 0.000 \\
motion\_blur s2  & 0.107 & \textbf{0.002} & 0.017 & 0.010 \\
motion\_blur s3  & 0.047 & 0.058 & 0.065 & 0.068 \\
resample s1      & 0.090 & \textbf{0.002} & 0.018 & 0.010 \\
resample s2      & 0.067 & \textbf{0.038} & 0.049 & 0.049 \\
resample s3      & 0.036 & 0.069 & 0.075 & 0.079 \\
rotation s1      & 0.099 & \textbf{0.012} & 0.010 & 0.000 \\
rotation s2      & 0.097 & \textbf{0.008} & 0.025 & 0.019 \\
rotation s3      & 0.090 & \textbf{0.002} & 0.018 & 0.010 \\
\bottomrule
\end{tabular}
\end{table}

\section{Prompt templates}
\label{app:prompts}
All templates are reproduced verbatim. \texttt{\{question\}} and
\texttt{\{options\}} are substituted at runtime; options are presented as a
comma-separated list.

\paragraph{Qwen2-VL-2B-Instruct (selected, 5/5 parse rate).}
A few-shot template with an explicit worked example:

\begin{verbatim}
You must reply in exactly two lines.

Example reply:
Answer: pizza
Confidence: 85

Now do the same for this image.
Question: {question}
Options: {options}
\end{verbatim}

\paragraph{Qwen2-VL rejected candidates.} Two alternatives were piloted on the
same five items and discarded. Template A (0/5 parse rate) returned bare
numbers such as \texttt{100} with no answer field:

\begin{verbatim}
Question: {question}
Options: {options}

Answer the question, then rate your confidence from 0 to 100.
Use exactly this format and nothing else:
Answer: <option>
Confidence: <number>
\end{verbatim}

Template C (0/5) returned bare labels such as \texttt{Takoyaki} with no
confidence field:

\begin{verbatim}
{question}
Choose one: {options}

Reply with two lines only:
Answer: (your choice)
Confidence: (0-100)
\end{verbatim}

\paragraph{SmolVLM-Instruct candidates (best 1/5).} All three templates below
were piloted; the model returned bare answers (``Takoyaki.'', ``Cheese
plate.'', ``Pho.'') without a confidence field in all but one of fifteen
attempts.

\begin{verbatim}
(A)  {question}
     Options: {options}

     Answer: <one option>
     Confidence: <0-100>

(B)  Look at the image.
     {question}
     Options: {options}

     Give your answer and how sure you are.
     Answer: pizza
     Confidence: 85

     Now yours:
     Answer:

(C)  {question}
     Choose from: {options}
     Then on a NEW line write Confidence: followed by a number 0-100.
     Answer:
\end{verbatim}

\paragraph{SmolVLM final template (internal confidence only).} Because no
template reliably elicited verbalized confidence, SmolVLM was run with a
minimal prompt and evaluated on internal token probability alone:

\begin{verbatim}
{question}
Options: {options}
Answer:
\end{verbatim}

\paragraph{Parsing.} Answers were extracted with the regular expression
\texttt{Answer:\textbackslash s*(.+)} and confidences with
\texttt{Confidence:\textbackslash s*([0-9]\{1,3\})}, the latter normalised by
dividing by 100. Responses without a parseable confidence were recorded and
counted toward the per-condition parse-failure rate reported in
Section~\ref{sec:results}.

\end{document}